\documentclass[10pt,twocolumn,letterpaper]{article}

\usepackage[pagenumbers]{cvpr} 
\usepackage[dvipsnames]{xcolor}

\usepackage{makecell}
\usepackage{bm}
\renewcommand{\vec}[1]{\bm{#1}}
\newcommand{\xx}{\vec{x}}
\newcommand{\xxb}{\vec{\bar{x}}}
\newcommand{\cc}{\vec{c}}
\newcommand{\ccb}{\vec{\bar{c}}}
\newcommand{\ee}{\vec{e}}
\newcommand{\reals}{\mathbb{R}}
\newcommand{\CC}{\mathcal{C}}
\renewcommand{\enspace}{}

\usepackage{multirow}
\usepackage{arydshln}  % For \cdashline

\newcommand{\best}[1]{{\bfseries{#1}}}
\newcommand{\tabnum}[2]{$#1\textcolor{gray}{\scriptstyle \pm #2}$}

\usepackage{overpic}
\usepackage{contour}

\usepackage{siunitx} % For number formatting using \num

\definecolor{cvprblue}{rgb}{0.21,0.49,0.74}
\usepackage[pagebackref,breaklinks,colorlinks,citecolor=cvprblue]{hyperref}

\def\paperID{12747}
\def\confName{CVPR}
\def\confYear{2025}

\title{Dense Match Summarization for Faster Two-view Estimation}

\author{Jonathan Astermark, Anders Heyden, and Viktor Larsson\\
Centre for Mathematical Sciences, Lund University\\
{\tt\small \{jonathan.astermark, anders.heyden, viktor.larsson\}@math.lth.se}
}

\begin{document}
\maketitle
\begin{abstract}
In this paper, we speed up robust two-view relative pose from dense correspondences. 
Previous work has shown that dense matchers can significantly improve both accuracy and robustness in the resulting pose. However, the large number of matches comes with a significantly increased runtime during robust estimation in RANSAC. To avoid this, we propose an efficient match summarization scheme which provides comparable accuracy to using the full set of dense matches, while having 10-100x faster runtime.
We validate our approach on standard benchmark datasets together with multiple state-of-the-art dense matchers.

\end{abstract}    
\section{Introduction}
\label{sec:intro}
Determining the two-view camera geometry is an important sub-task in many computer vision problems, \eg~for Simultaneous Localization and Mapping or Structure-from-Motion.
The most common approach is to first detect corresponding points in the two views, followed by robust estimation using RANdom Sample Consensus (RANSAC)~\cite{fischler1981ransac} which both estimates the relative pose and identifies potential outlier matches.
Traditionally, keypoint matching is performed by independently detecting keypoints in each view, followed by a matching step, either directly comparing descriptor similarity or using a learned matcher.

Recently, a lot of attention has been given to so-called detector-free matching, which takes the images as input and directly establishes (semi-)dense correspondences.
These matchers produce significantly more matches, compared to the detector-based counterparts, especially in weakly textured image regions, such as walls, floors, and ceilings.
Dense matching shows great promise 
and currently achieves the most accurate two-view estimates in standard benchmarks.
However, this comes with a trade-off in runtime, both for running the matching and in the subsequent robust estimation (\ie RANSAC).

\begin{figure}[t]
    \begin{overpic}[width=\columnwidth, trim={0 0.5cm 4cm 0}, clip]{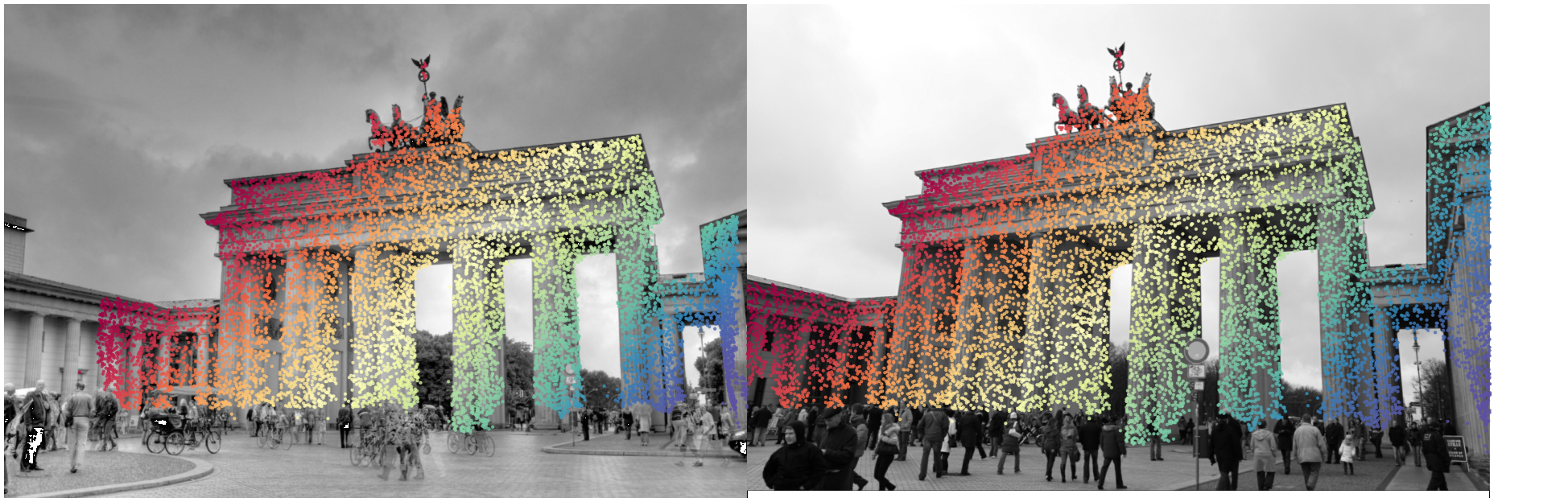}
    \put(2.5,29){\textcolor{white}{$\epsilon = 0.38^\circ$}}
    \put(2.5,24){\textcolor{white}{$t = 98$ ms}}
    \end{overpic}
    \\[0.1cm]
    \begin{overpic}[width=\columnwidth, trim={0 0.5cm 4cm 0}, clip]{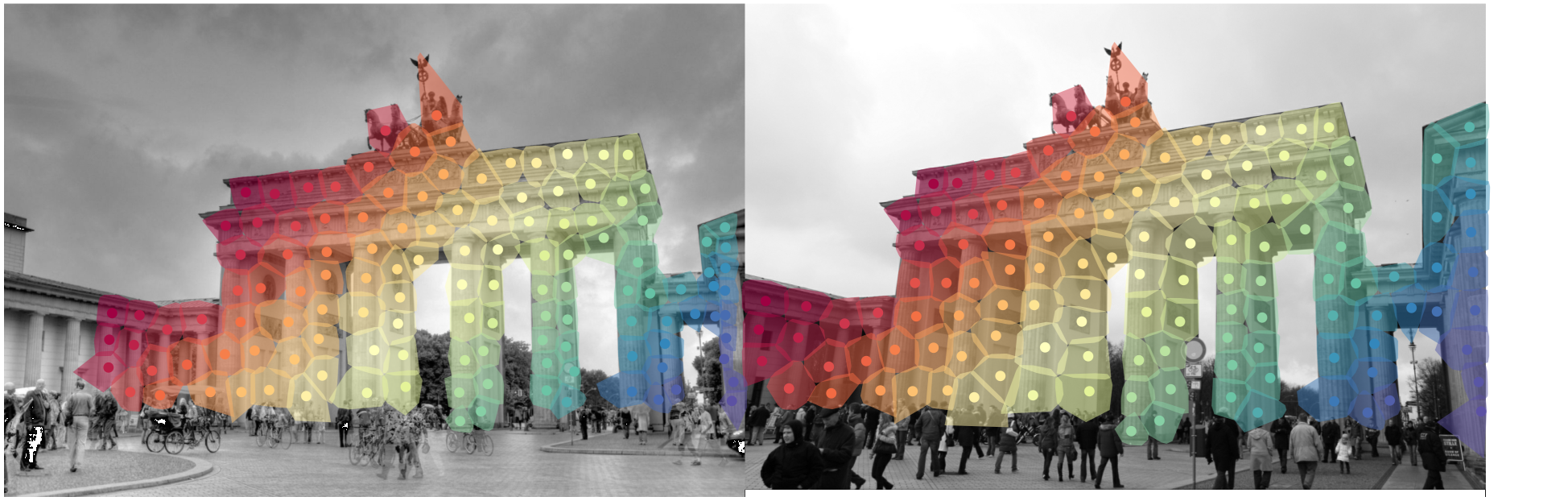}
    \put(2.5,29){\textcolor{white}{$\epsilon = 0.34^\circ$}}
    \put(2.5,24){\textcolor{white}{$\bm{t = 2.3}$ \textbf{ms}}}
    \end{overpic}
    
    \caption{\textbf{Relative Pose Estimation from Dense Matches.} The top images show 10,000 semi-dense matches from DKM~\cite{edstedt2023dkm} with same colors indicating corresponding points.
    The large number of matches makes robust estimation with RANSAC accurate (pose error $\epsilon = 0.38^\circ$) but slow (runtime $t = 98$ ms). 
    In this paper we show that we can get comparable accuracy with significantly lower runtime by sparsifying the dense matches (bottom images).
    Each sparsified match is represented by a $9 \! \times \! 9$-matrix that summarizes the geometric constraints from nearby matches.
    }\label{fig:teaser}
\end{figure}

In this paper, we aim to improve the runtime of the robust estimation step.
We show that when using dense matchers, most of the matches provide redundant geometric constraints, and we present an efficient match summarization scheme that selects a subset of $\approx \! 1 \%$ of the matches, which we call \textit{representative matches}, that are used for robust estimation.
Using this smaller set of matches, we can obtain a 10-100x speedup with a marginal loss in pose accuracy.
In addition, we present an approximation approach for enriching each representative match to capture the geometric constraints of nearby correspondences in a single $9 \! \times \! 9$-matrix (regardless of how many neighboring matches are included).
This enrichment allows for an even smaller loss of pose accuracy, while still being significantly faster compared to dense correspondence. In \cref{fig:teaser} we visualize one example from MegaDepth~\cite{Li2018megadepth}.

We validate our claims and our approach with extensive experiments and ablation studies on standard benchmarks with different state-of-the-art dense matchers.

In summary, our contributions are:
\begin{itemize}
    \item A scheme for summarizing any dense matches into sparse \textit{representative matches}, which speeds up the robust estimation by a factor 10-100x, with only a very small loss in accuracy. 
    \item An approximation-based approach to summarize the contributions of each cluster, which we use to refine the output from RANSAC. This lets us maintain the high accuracy from dense matches, with only a small increase in runtime compared to the approach above.
    \item In experiments on standard benchmarks, with multiple state-of-the-art dense matchers, we are able to consistently show a significant speedup.
    Furthermore, our experiments show that sparsified dense matches outperform state-of-the-art sparse matches in terms of accuracy.
\end{itemize}

\subsection{Related Work}

\noindent
\textbf{Matching and Correspondences.}
Image correspondence for geometric estimation traditionally works with a \textit{detect-then-match} paradigm that first identifies keypoints (\eg~SIFT~\cite{lowe2004sift}, SuperPoint~\cite{detone2018superpoint}) independently in each image, followed by a matching step.
The matching is either done by comparing descriptor similarity (\cite{lowe2004sift,detone2018superpoint,dusmanu2019d2}) or more recently via learned matchers (\cite{sarlin2020superglue, lindenberger2023lightglue}).
In~\cite{sun2021loftr} the authors instead proposed to perform \textit{detection-free} matching, which takes an image pair as input and directly regresses semi-dense pixel correspondences.
This started a series of work on dense matching (\cite{jiang2021cotr, edstedt2023dkm, chen2022aspanformer, edstedt2024roma}) which improved on the initial work.
Common for all dense methods is that they produce significantly more matches compared to classical detector-based methods.
By considering both images jointly, the methods are able to propagate strong matches to weaker ones, allowing for even weakly textured regions to be matched between images.

In benchmarks for two-view relative pose estimation, these methods significantly improve the results in terms of pose accuracy.
However, this comes at a trade-off with runtime, as the large number of matches makes robust estimation more costly.
In this work, we offset some of this cost by speeding up the robust estimation.
Our method is not specific to a particular matcher and can be applied to any dense correspondences.

\noindent
\textbf{Robust Estimation.}
For robust geometric estimation problems in computer vision, the RANSAC algorithm, originally introduced by Fischler and Bolles~\cite{fischler1981ransac} in 1981, is the de facto standard.
Modern RANSAC variants, while still working with similar principles as the original, usually incorporate some form of local optimization (LO-RANSAC~\cite{chum2003loransac,lebeda2012fixing}) as well a more complex scoring beyond simple inlier counting, \eg~MSAC~\cite{torrMLESACNewRobust2000} or MAGSAC~\cite{barath2019magsac}.
Other works improve by integrating various learned components in the pipeline, see \eg~\cite{cavalli2023consensus, wei2023adaptive, barroso2023two}.
There are also several works which focus on making RANSAC faster.
In~\cite{chum2008optimal}, the authors propose a probabilistic method (SPRT) for deciding when to early exit.
In~\cite{korman2018latent}, the authors propose to only perform scoring once two similar models are found.
Rais et al.~\cite{rais2017accurate} instead aggregate multiple model hypotheses to form the final output.
Barath et al.~\cite{barath2022space} speed up scoring by filtering out regions of the matches which cannot contain inlier correspondences for the current model.
PROSAC \cite{chum2005prosac} leverages per-match confidences to sample good models earlier, to speed up convergence.
Ni et al.~\cite{ni2009groupsac} group correspondences and model different inlier ratios for each cluster, which is used both for sampling and stopping criterion.

The match summarization method proposed in this paper is orthogonal (and could be combined) with the improvements in the above methods.
In particular, most prior work focus on either speeding up the scoring or reducing the number of iterations, but do not consider refinement which is particularly costly with a large number of matches.

\noindent
\textbf{Richer Correspondences.}
For \textit{affine-correspondences} (ACs) each match is associated with a $2\times 2$ matrix representing a local transformation around the keypoints.
These matrices can be interpreted as a linearization of the local planar homography~\cite{raposo2016theory}.
Affine correspondences have been used to derive more efficient minimal estimators (\cite{bentolila2014conic,raposo2016theory,eichhardt2018affine,ventura2023p1ac}) as well as provide constraints for normal estimation~\cite{hajder2023fast}.
In this paper, we derive a summarized correspondence expressed with a $9\times 9$-matrix. 
While this matrix is similar to ACs in that it encodes additional local geometric constraints, it does not make assumption on local planarity and yields stronger geometric constraints, even allowing estimation from a single correspondence (see \cref{sec:ablation_ransac}).

\section{Background}
\label{sec:background}

Each 2D-point correspondence $(\xx,\xxb) \! \in \! \reals^3 \times \reals^3$ (in homogeneous coordinates) constrain the relative pose by
\begin{equation}\label{eq:epipolar}
    \xxb^T E \xx = 0 \enspace ,
\end{equation}
where $E \! = \! [\vec{t}]_\times R$ is the essential matrix.
As the points $(\xx,\xxb)$ are measurements from the images, they will contain noise and will not satisfy \eqref{eq:epipolar} exactly, even if the match is correct.
Thus, in the optimization, a residual such as the Sampson error is commonly used instead,
\begin{equation}\label{eq:sampson_original}
    \mathcal{E}(E,\xx,\xxb) = \frac{(\xxb^T E \xx)^2}{\|E_{12}\xx\|^2 + \|(E^T)_{12}\xxb\|^2},
\end{equation}
where $E_{12}$ is the first two rows of $E$.
The Sampson error approximates the squared reprojection error, and is used to determine match correctness (inlier or outlier).
Most RANSAC-variants additionally use some form of MSAC-scoring~\cite{torrMLESACNewRobust2000} for the models, \ie~the essential matrix is chosen by minimizing the sum of truncated residuals
\begin{equation} \label{eq:original_cost}
    f(E) = \sum_{i = 1}^N \min \left\{ \mathcal{E}(E,\xx_i, \xxb_i), \tau^2 \right\} \enspace ,
\end{equation}
where $N$ denotes the number of matches and $\tau \! \in \! \reals_+$  is the inlier threshold.
The model-scoring function $f(E)$ is used both for selecting the best model, and for  non-linear refinement.
As the number of matches $N$ grows large, evaluating $f(E)$ will dominate the runtime cost in RANSAC.

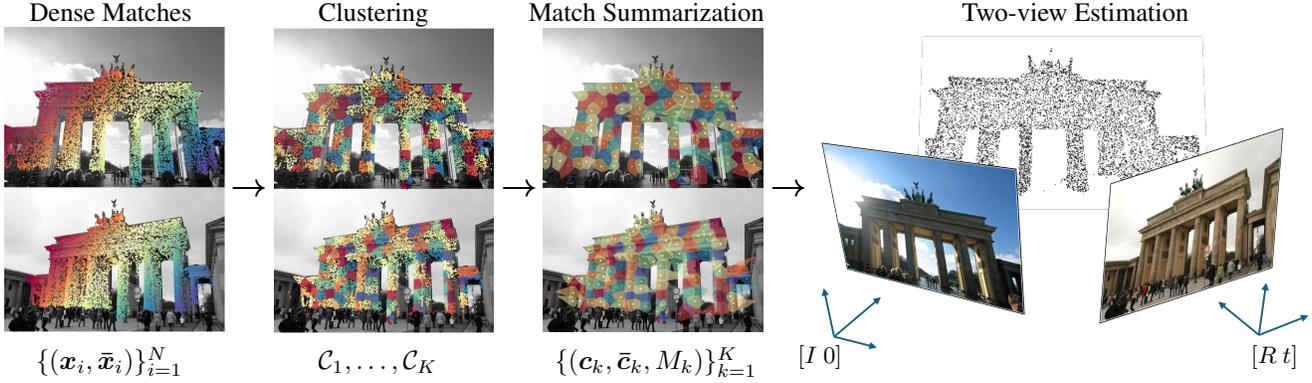
\begin{figure*}[t]
    \centering
    \begin{overpic}[width=\textwidth]{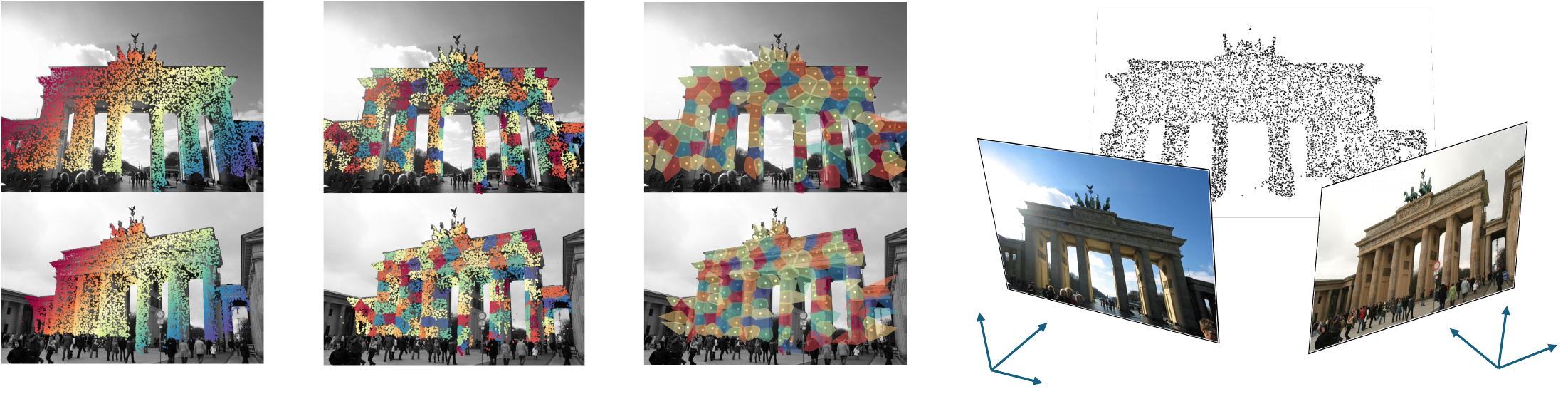}
    \put(2,25.5) {Dense Matches}
    \put(2.5,-1.0) {$\{(\xx_i,\xxb_i)\}_{i=1}^N$}    
    \put(24,25.5) {Clustering}
    \put(24,-1.0) {$\CC_1,\dots,\CC_K$}    
    \put(40,25.5) {Match Summarization}
    \put(42,-1.0) {$\{(\cc_k,\ccb_k, M_k)\}_{k=1}^K$}
    \put(73,25.5) {Two-view Estimation}
    \put(60.5,-0.5) {\small $[I~0]$}
    \put(95,-0.5) {\small $[R~t]$}
    \put(17.3,12.2) {\large $\boldsymbol{\rightarrow}$}
    \put(37.9,12.2) {\large $\boldsymbol{\rightarrow}$}
    \put(58.4,12.2) {\large $\boldsymbol{\rightarrow}$}

    \end{overpic}
    \\[0.2cm]
    \caption{\textbf{Overview of our Summarization Scheme.} As input, our method takes a set of \textbf{dense matches} $\{(\xx_i,\xxb_i)\}_{i=1}^N$, where typically $N = \num{10000}$. Through \textbf{clustering}, the matches are grouped into clusters $\CC_1,\dots,\CC_K$, $K \ll N$, which yield approximately the same geometric constraints on the relative pose. For each cluster, we then perform \textbf{match summarization}, which replaces each group of matches with a representative match $(\cc_k,\ccb_k)$ and a $9\times 9$-matrix $M_k$ that encodes the geometric constraints. These summarized meta-matches can then used for robust \textbf{two-view estimation} with significant speedup for a very small accuracy loss.}
    \label{fig:overview}
\end{figure*}

\section{Method}\label{sec:method}
In this section, we present a method for efficiently summarizing the geometric constraints given by the dense matches.
The idea is to first cluster the correspondences into subsets that yield similar geometric constraints, and replace each cluster with a single \textit{representative match} used in robust estimation.
Next, we compute a \textit{proxy residual} that summarizes the geometric constraints from each cluster into a $9\times 9$-matrix, irrespective of  cluster size. This allows us to refine the pose without evaluating the full residual~\eqref{eq:original_cost} in each step of the optimization. \cref{fig:overview} shows an overview of our method.
In~\cref{sec:clustering} we discuss different approaches for performing the clustering and selecting representative matches, and in~\cref{sec:approx} we derive the proxy residual. 

In the paper we focus the presentation on the calibrated case (essential matrix), but the approach is directly applicable to the uncalibrated (fundamental matrix) setting as well.

\subsection{Clustering and Representative Matches} \label{sec:clustering}
Given a set of dense matches, our goal is to find a sparse subset of representative matches
that capture the same geometric constraints as the full set.
To do this, we begin by clustering correspondences that contribute with similar residuals.
From each cluster, we then select a single match which serves as a representative of that cluster.

We base our clustering on the assumption that matches close to each other should lead to similar residuals in~\eqref{eq:original_cost}. 
This motivates us to group matches based on their position in the two images.
A simple approach is to use a standard clustering method, \eg K-means, applied on the 4-dimensional match vectors (\ie concatenated keypoint coordinates from both images).
Alternatively, matches could be clustered based on the images, \eg using superpixels.
In~\cref{sec:ablation_clustering}, we discuss and compare different approaches for clustering and their trade-off in terms of runtime and accuracy for our method.

For each cluster, we then select the match closest to the cluster centroid (in 4D match space) as the representative match. 
The set of representative matches $\{(\cc_i,\ccb_i)\}_{i=1}^K \subset \{(\xx_i,\xxb_i)\}_{i=1}^N$, where K is the number of clusters, can be used directly in RANSAC to estimate the relative pose, using the sub-sampled cost
\begin{equation}\label{eq:cost_cluster}
    f_c (E) = \sum_{i=1}^K \min \left\{ \mathcal{E}(E, \cc_i, \ccb_i), \tau^2 \right\} \enspace .
\end{equation}
Since the runtime of RANSAC is generally dominated by scoring and refinement, which grows linearly with the number of matches, significant subsampling can drastically reduce the runtime.
In our experiments (Section~\ref{sec:experiments}) we show that even when severely subsampling ($K \! \approx \! N/80$), leading to 55x speedup, we can still obtain good relative pose estimates when using this method.
This highlights that most constraints in the dense matches are redundant.
However, the sparsified matches still will lead to a slight drop in accuracy.
In the next section, we propose an approximation which enriches each of the matches to also encode the geometric constraints from the full cluster.

\subsection{Dense Match Summarization}\label{sec:approx}

We now present a method for approximating the full cost in \eqref{eq:original_cost} using the clustering.
The idea is to replace the dense matches in each cluster with a more computationally efficient proxy residual.
While we in the previous section reduced the number of residuals from $N$ to $K$ by only considering the representative matches, in this section we will capture more of the geometry per cluster by increasing this number to $9K$. We do this in two steps; first by assuming that each cluster is either all-inlier or all-outler, and second by approximating the Sampson error in each cluster.

Let $\mathcal{M} \! \subset \! \reals^3\times \! \reals^3$ be the set of dense input matches, and let $\CC_1,\dots,\CC_K \! \subset \! \mathcal{M}$ be a disjoint clustering of these with associated representative matches $\{(\cc_k,\ccb_k)\}_{k=1}^K$. 
By assuming that the matches in each cluster are either all inliers or all outliers,~\eqref{eq:original_cost} can be rewritten as
\begin{align}
    f(E) &= \sum_{k=1}^K \sum_{(\xx,\xxb)\in \CC_k} \min \left\{ \mathcal{E}(E,\xx,\xxb),\tau^2\right\} \\
    &\approx \sum_{k=1}^K \min\left\{ \sum_{(\xx,\xxb)\in \CC_k} \mathcal{E}(E,\xx,\xxb),~|\CC_k|\tau^2\right\} \label{eq:approx_cluster_inlier_outlier} \enspace,
\end{align}
where $~|\CC_k|$ denotes the number of matches in cluster $\CC_k$.
This means that each cluster either contributes the sum of all its dense residuals, or the constant factor  $~|\CC_k|\tau^2$  -- whichever is smallest.

Now, consider a single cluster $\CC$ with representative match $(\cc,\ccb)$. The sum of all its dense residuals is
\begin{equation}
    f_{cl}(E; \CC) = \sum_{(\xx,\xxb)\in \CC} \mathcal{E}(E,\xx,\xxb) \enspace ,
\end{equation}
\ie\ the inner sum in~\eqref{eq:approx_cluster_inlier_outlier}.
By construction, each match $(\xx,\xxb) \! \in \! \CC$ will lie relatively close to the cluster's representative match $(\cc,\ccb)$. This motivates us to approximate the Sampson error~\eqref{eq:sampson_original} for each dense match as
\begin{align}
    \mathcal{E}(E,\xx,\xxb) \approx \frac{(\xxb^T E \xx)^2}{\|E_{12}\cc\|^2 + \|(E^T)_{12}\ccb\|^2} \enspace ,
    \label{eq:approx}
\end{align}
\ie by replacing $(\xx,\xxb)$ in the denominator with $(\cc,\ccb)$.
For brevity, we introduce  $\alpha(E; \CC) = \|E_{12}\cc\|^2 + \|(E^T)_{12}\ccb\|^2$. Then the approximation in \eqref{eq:approx} lets us write the sum of residuals for a cluster more compactly as
\begin{align}
     f_{\text{cl}}(E; \CC)
     &\approx
     \sum_{(\xx,\xxb)\in\CC} \frac{(\xxb^TE\xx)^2}{\alpha(E; \CC)} \\
     &=
     \frac{1}{\alpha(E; \CC)} \left\|
    \begin{pmatrix}
    \xxb_1^T E \xx_1 \\ \vdots \\ \xxb_n^T E \xx_n
    \end{pmatrix}\right\|^2 
    = 
    \frac{1}{\alpha(E; \CC)} \|A \ee \|^2 \enspace ,
    \label{eq:cost_cluster_approx}
\end{align}
where $n = |\CC_k|$; $\ee \in \reals^{9}$ is a vector containing the elements of $E$,
and $A \in \reals^{n \times 9}$ is the matrix with rows $A_i = (\xx_i \otimes \xxb_i)^T$ for all $(\xx_i,\xxb_i) \in \CC$. Here, $\otimes$ denotes the Kronecker product.
The trick to efficiently evaluating~\eqref{eq:cost_cluster_approx} is that the large matrix $A$ can be replaced with the \textit{reduced measurement matrix} \cite{rodriguez_reduced_2011} $M \in \reals^{9\times 9}$ using Cholesky factorization $A^TA = M^TM$, since
\begin{equation}
      \|A\vec{e}\|^2 =\vec{e}^TA^TA\vec{e} = \vec{e}^TM^TM\vec{e} = \|M\vec{e}\| ^2 \enspace .
\end{equation}
Note that this is very fast to compute (in our experiments a total of 0.2 ms for 128 clusters with $N = 10,000$) and only needs to be done once for each clustering.

The final approximate cost function for all matches, to be evaluated in each step of the optimization, is
\begin{equation}\label{eq:cost_approx}
    f_{approx}(E) =
    \sum_{k=1}^K \min \left\{
        \frac{1}{\alpha(E; \CC_k)} \left\| M_k \ee \right\|^2,
        |\CC_k| \tau^2
    \right\} \enspace .
\end{equation}
In our experiments (\cref{sec:evaluate_approx}) we show that this provides a tight approximation to the sum of Sampson errors, while being significantly faster to compute.

Above, the approximation was applied to the Sampson residual, but it is in principle also applicable to residuals with similar form, \eg~Symmetric Epipolar Distance:
\begin{equation} \nonumber
     \sum_{(\xx,\xxb)} \left(\frac{(\xxb^TE\xx)^2}{\|E_{12}\xx\|^2} + \frac{(\xxb^TE\xx)^2}{\|E^T_{12}\xxb\|^2} \right) \approx 
        \frac{\|A\boldsymbol{e}\|^2}{\|E_{12}\cc\|^2} + \frac{\|A\boldsymbol{e}\|^2}{\|E^T_{12}\ccb\|^2}.
\end{equation}

\section{Experiments}
\label{sec:experiments}
In this section, we evaluate the match clustering and summarization scheme presented in~\cref{sec:method}, through experiments on real image pairs.
First, in~\cref{sec:ablation_clustering}, we compare clustering methods and perform an ablation study on both the clustering method and number of clusters. In~\cref{sec:evaluate_approx}, this is followed by an evaluation of the approximation error of our summarization scheme.
Next, in~\cref{sec:ablation_ransac}, we perform an ablation study on the integration of our summarization scheme in RANSAC.

We use two standard benchmarks for relative pose estimation, ScanNet-1500~\cite{sarlin2020superglue, dai2017scannet} and MegaDepth-1500~\cite{sun2021loftr, Li2018megadepth}, which each contain 1500 image pairs of indoor and outdoor scenes, respectively.
For inlier thresholds, we use 1.0 pixels in MegaDepth-1500, and 2.5 pixels in ScanNet-1500.
Following prior work, we report the pose error (max of rotation and translation error in degrees) with the Area Under Curve (AUC) up to some threshold.
For our ablations and approximation evaluation, we evaluate on MegaDepth-1500, using $N=\num{10000}$ dense matches from DKM~\cite{edstedt2023dkm}. Finally, in \cref{sec:final_comparison} we show that our results generalize to other state-of-the-art dense matchers and other datasets. 
We also present results on estimation of the fundamental matrix on the challenging WxBS dataset~\cite{mishkin2015wxbs}.

\subsection{Implementation Details}

We implement our approximate residual in an LO-RANSAC framework building on PoseLib~\cite{poselib}, which is a state-of-the-art robust estimation library in C++.
To get a fair comparison with the dense and subsampled methods, we implement the standard Sampson residual in the same framework.
For the ablations and evaluations in the following sections, our method runs RANSAC using the representative matches from~\cref{sec:clustering}, followed by refinement using the approximate residuals described in~\cref{sec:approx}.
All timings are measured on a modern desktop CPU.

\subsection{Ablation on Clustering}\label{sec:ablation_clustering}
First, we explore different approaches for performing the match clustering.
We explore both keypoint-based and image-based clustering methods. 
For the keypoint-based methods, we evaluate K-means clustering on both the 4-dimensional match vectors and the 2-dimensional keypoints from one of the images.
Motivated by the geometric constraints in~\eqref{eq:cost_cluster_approx},
we also evaluate clustering in the 9-dimensional constraint vectors $A_i = \xx_i \otimes \xxb_i \in \mathbb{R}^{9}$.
For the K-means clustering experiments we use FAISS~\cite{douze2024faiss} on a CPU, running a maximum of 5 iterations.

For image-based clustering, we first segment one of the images, then consider all matches within the same segment as a cluster. We evaluate both a simple 2D-grid segmentation, and a more sophisticated superpixel detection.
The 2D grid is created by dividing the image region between the minimum and maximum keypoint coordinates into $m^2$ equally sized rectangles, where $m=\lceil \sqrt{K} \rceil$. 
For the superpixel detection, we use SLIC~\cite{achanta2012slic, algy2024fastslic}. 
For each clustering method, we select the representative match 
by finding the match closest to the centroid.

\begin{figure*}
        \centering
     \begin{overpic}[width=0.32\textwidth, trim={0 0.1cm 1cm 0}, clip]{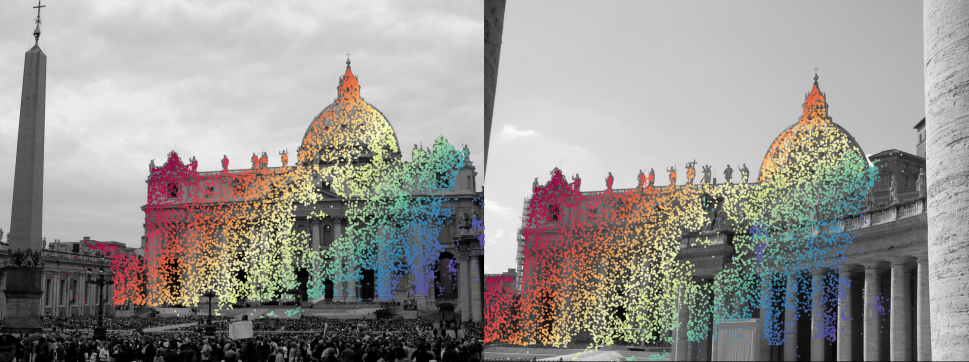}
      \put (0,32){\small \colorbox{white}{Input (dense)}}
     \end{overpic}
     \begin{overpic}[width=0.32\textwidth, trim={0 0.1cm 1cm 0}, clip]{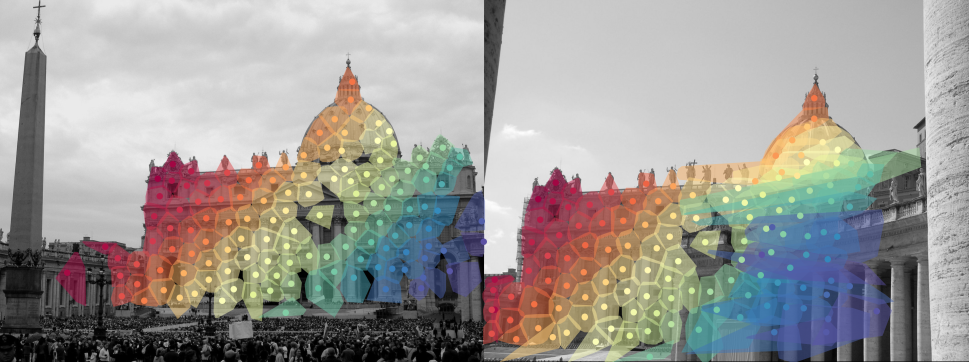}
      \put (0,32){\small \colorbox{white}{2D}}
     \end{overpic}
     \begin{overpic}[width=0.32\textwidth, trim={0 0.1cm 1cm 0}, clip]{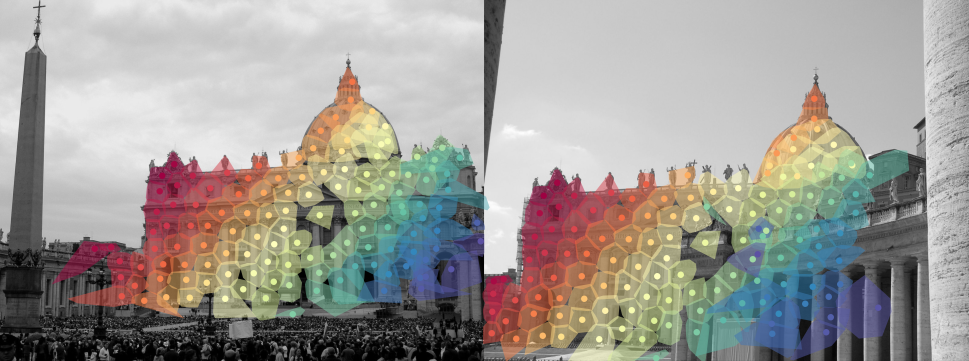}
      \put (0,32){\small \colorbox{white}{4D}}
     \end{overpic}
     \begin{overpic}[width=0.32\textwidth, trim={0 0.1cm 1cm 0}, clip]{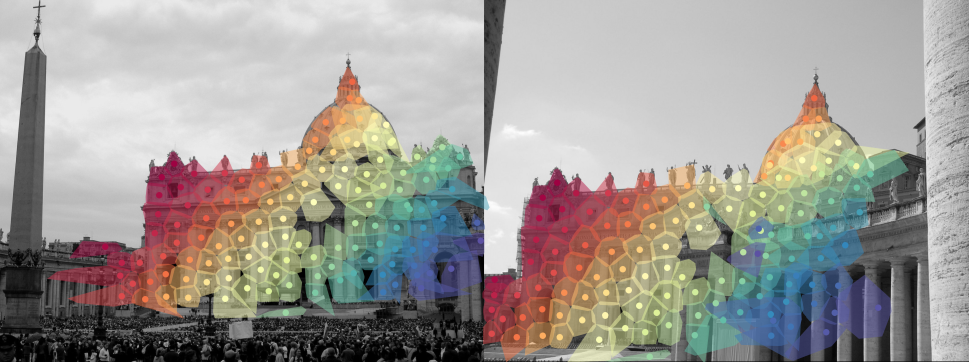}
      \put (0,32){\small \colorbox{white}{9D}}
     \end{overpic}
     \begin{overpic}[width=0.32\textwidth, trim={0 0.1cm 1cm 0}, clip]{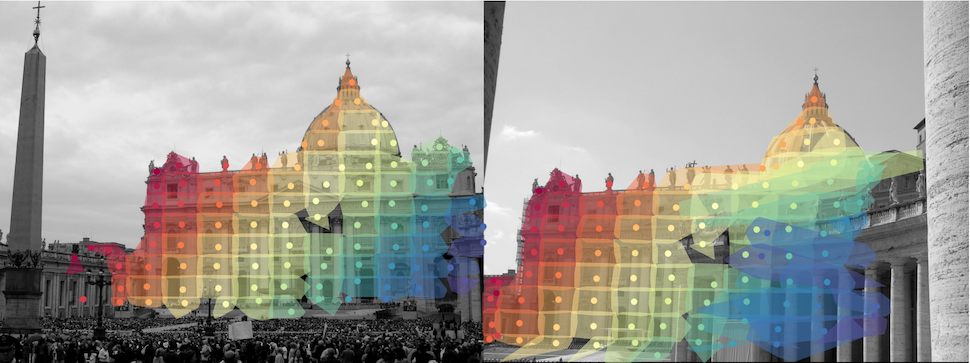}
      \put (0,32){\small \colorbox{white}{Grid}}
     \end{overpic}
     \begin{overpic}[width=0.32\textwidth, trim={0 0.1cm 1cm 0}, clip]{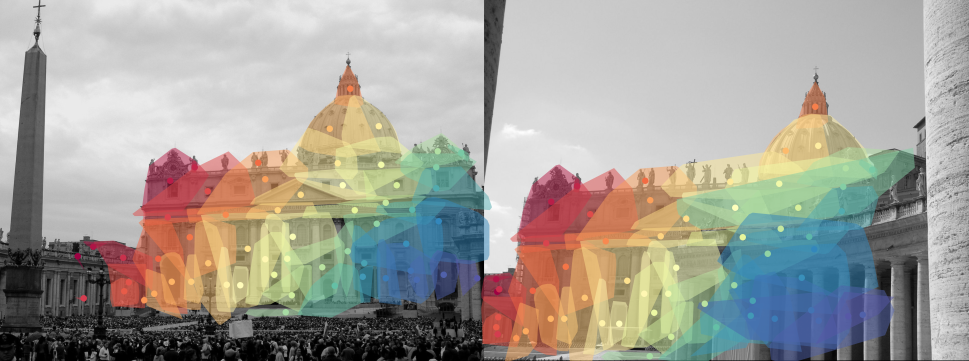}
      \put (0,32){\small \colorbox{white}{SLIC}}
     \end{overpic}
\\[0.2cm]
     \begin{overpic}[width=0.32\textwidth, trim={0 0.1cm 1cm 0}, clip]{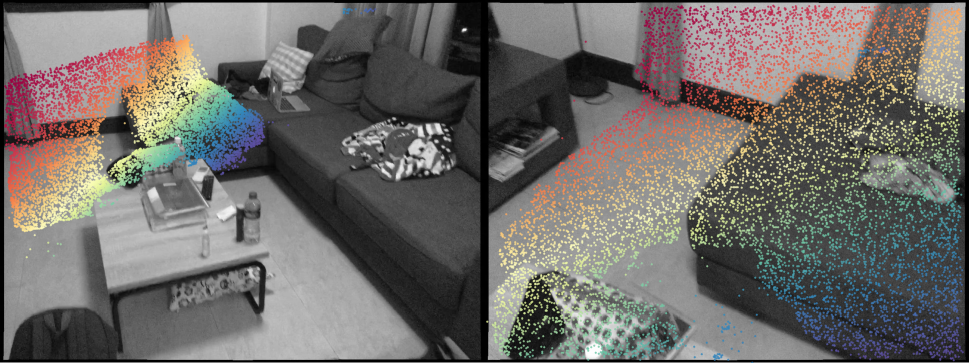}
      \put (0,2){\small \colorbox{white}{Input (dense)}}
     \end{overpic}
     \begin{overpic}[width=0.32\textwidth, trim={0 0.1cm 1cm 0}, clip]{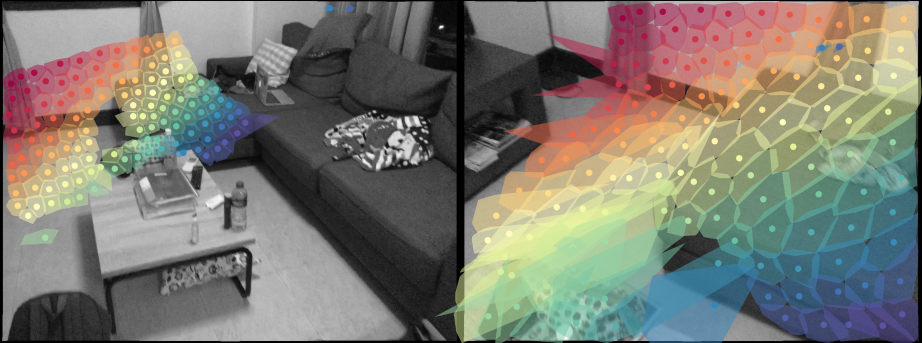}
      \put (0,2){\small \colorbox{white}{2D}}
     \end{overpic}
     \begin{overpic}[width=0.32\textwidth, trim={0 0.1cm 1cm 0}, clip]{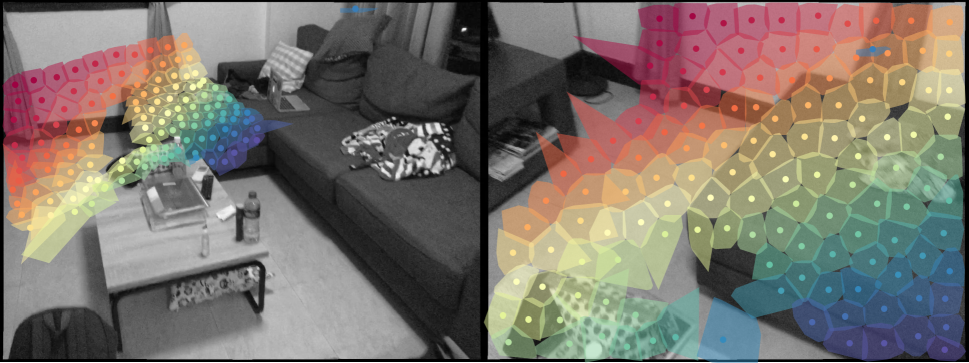}
      \put (0,2){\small \colorbox{white}{4D}}
     \end{overpic}
     \begin{overpic}[width=0.32\textwidth, trim={0 0.1cm 1cm 0}, clip]{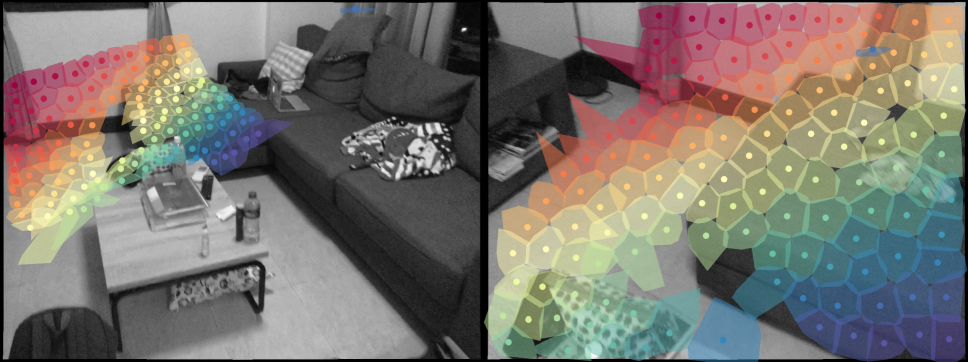}
      \put (0,2){\small \colorbox{white}{9D}}
     \end{overpic}
     \begin{overpic}[width=0.32\textwidth, trim={0 0.1cm 1cm 0}, clip]{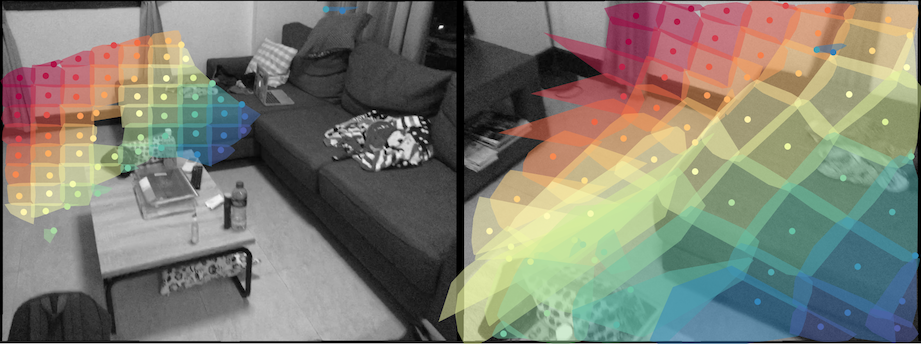}
      \put (0,2){\small \colorbox{white}{Grid}}
     \end{overpic}
     \begin{overpic}[width=0.32\textwidth, trim={0 0.1cm 1cm 0}, clip]{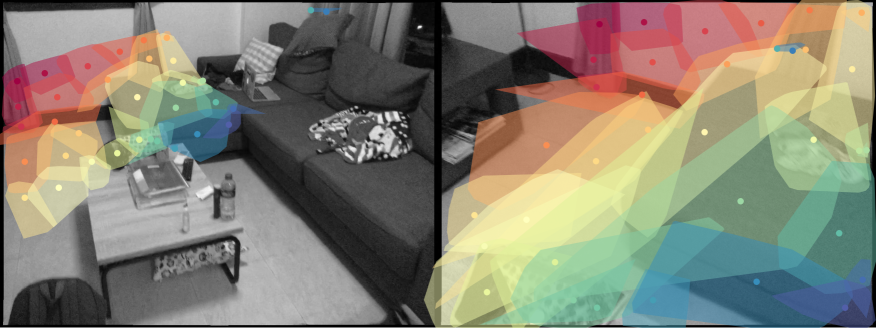}
      \put (0,2){\small \colorbox{white}{SLIC}}
     \end{overpic}
    \caption{\textbf{Qualitative Comparison of Clustering Methods.} We compare different clustering methods on a single image pair from MegaDepth-1500 (top 2 rows) and ScanNet-1500 (bottom 2 rows), respectively. Dense matches from DKM are clustered using three keypoint-based methods (2D, 3D, 9D) and two image-based methods (Grid, SLIC).}
    \label{fig:clustering_qualitative}
\end{figure*}

In Figure~\ref{fig:clustering_qualitative}, we show some qualitative examples of the different clusterings on two example image pairs from MegaDepth-1500 and ScanNet-1500, respectively. In the top example, we see that the keypoint-based methods (2D, 4D, 9D) all give similar clustering patterns, while in the bottom example 2D clustering leads to a coarser clustering in the second image. This is explained by the smaller image overlap; while 4D and 9D clustering takes the keypoint density in both images into account, 2D clustering only sees the keypoints from one of the images.
The image-based methods (Grid and SLIC), on the other hand, lead to larger clusters in both of the image pairs. This is because the segmentation is not proportional to the match density, so they may ``waste'' clusters on image regions with few or no matches.

In Table~\ref{tbl:clustering_ablation}, we compare the errors and runtimes for different clustering methods on MegaDepth-1500, using a cluster size of 128. We report both the time for clustering, and for estimation using the clusters. 
We assume that grid-clustering can be done at negligible runtime with an efficient implementation.
In the table, we see that the difference between clustering methods is quite small, except for SLIC which is markedly more expensive to calculate.
A key difference with image-based clustering, however, is that it only needs to be done once per image. Keypoint-based clustering, on the other hand, needs to be done once per image pair, which can lead to many more total evaluations. For large datasets with high co-visibility, this may be important to consider.

\begin{table}[ht]
\small
\centering
    \begin{tabular}{l c c r r} \toprule
           &               &                   & \multicolumn{2}{c}{Runtime} \\ \cmidrule{4-5}
    Method & AUC@5$^\circ$ & $\epsilon_{avg}$ & Clustering & RANSAC \\ \midrule
    K-means 2D & 66.4 & 3.62 & 1.53 ms & 1.46 ms  \\
    K-means 4D & 66.9 & 3.28 & 1.22 ms & 1.44 ms \\
    K-means 9D & 66.6 & 3.25 & 1.38 ms & 1.43 ms\\
    Grid & 65.9 & 3.73 & $\approx 0$ ms & 1.34 ms \\
    SLIC~\cite{algy2024fastslic} & 64.5 & 4.07 & 22.38 ms & 0.97 ms \\
    \bottomrule
    \end{tabular}
\caption{\textbf{Clustering Method Ablation.} The table shows the pose errors for each clustering approach on MegaDepth-1500, as well as the average runtime for each clustering strategy. }
\label{tbl:clustering_ablation}
\end{table}

\begin{figure*}
    \centering
    \resizebox{1.0\textwidth}{!}{
        \includegraphics[]{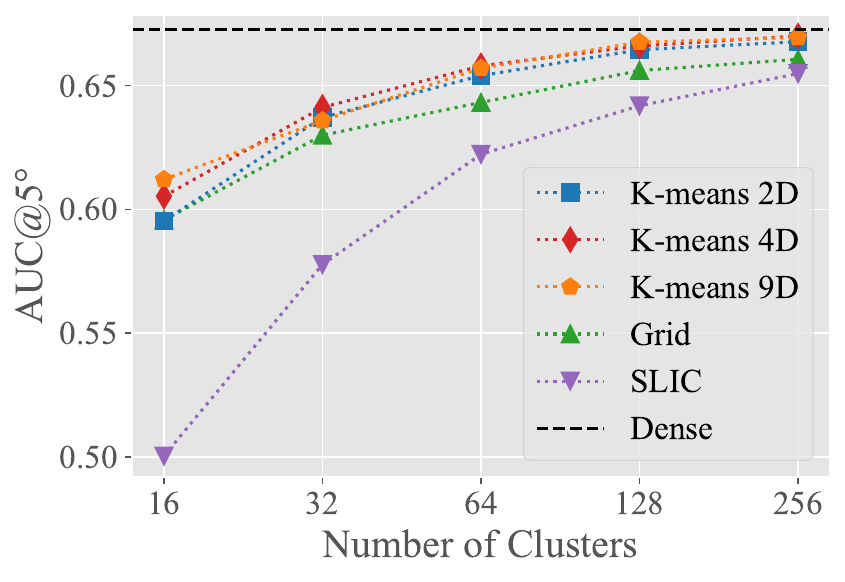}
        \includegraphics[]{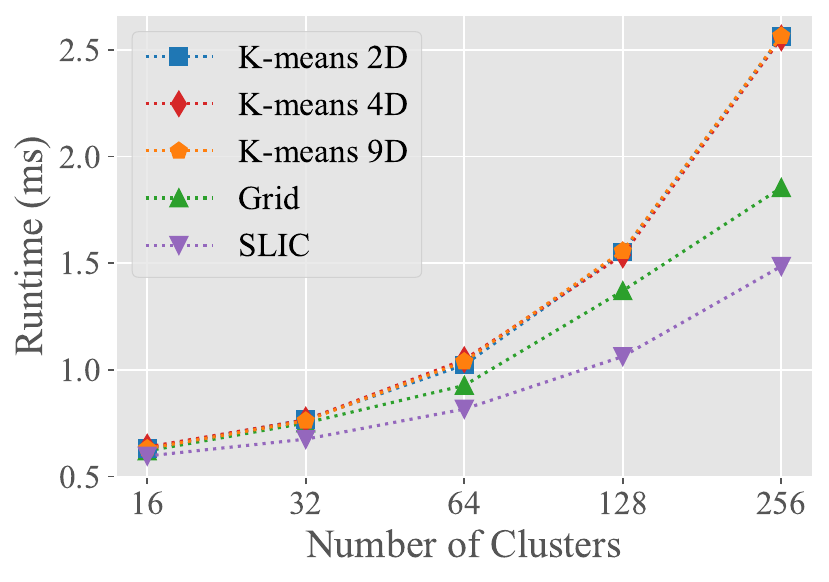}
        \includegraphics[]{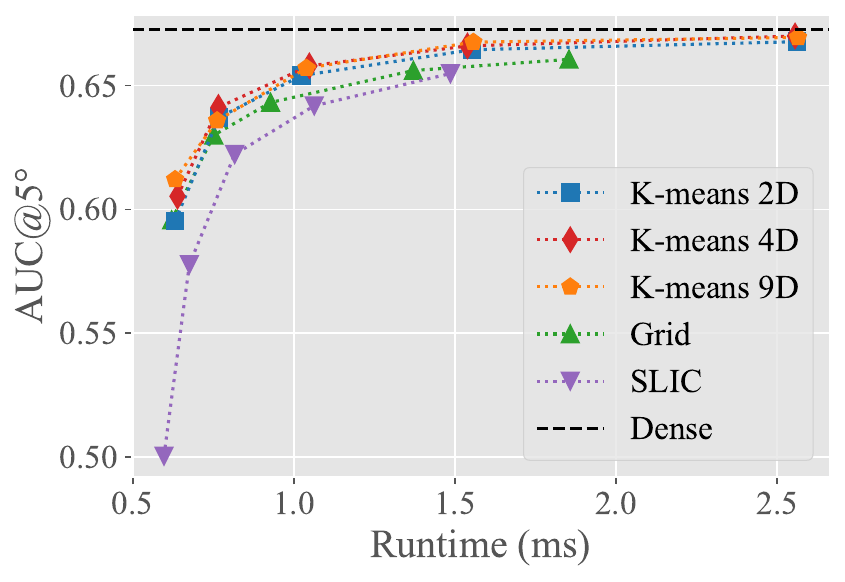}
    }
    \caption{\textbf{Clustering Method and Hyperparameter Ablation.} The plots show AUC@$5^\circ$ (left), runtime (middle) and AUC-runtime-tradeoff (right) for different clustering methods and number of clusters $K$. We include AUC@$5^\circ$ for the dense baseline as reference.}
    \label{fig:ablation_clustering}
\end{figure*}

Next, we evaluate the impact of the number of clusters.
Figure~\ref{fig:ablation_clustering} shows the AUC@$5^\circ$ (left) and runtime (center) plotted against the number of clusters, as well as AUC@$5^\circ$ vs. runtime (right). We also include the dense baseline (dashed).
The results further indicate that all K-means clustering variants perform similarly, and that the performance gets close to the dense baseline as number of clusters increases. 
Based on these results, we select K-means 4D with $K=128$ as our main clustering method to be used in the  experiments, but note that other choices are viable as well.

\subsection{Evaluation of Approximation Error}\label{sec:evaluate_approx}

\begin{figure}[ht]
    \includegraphics[width=\columnwidth]{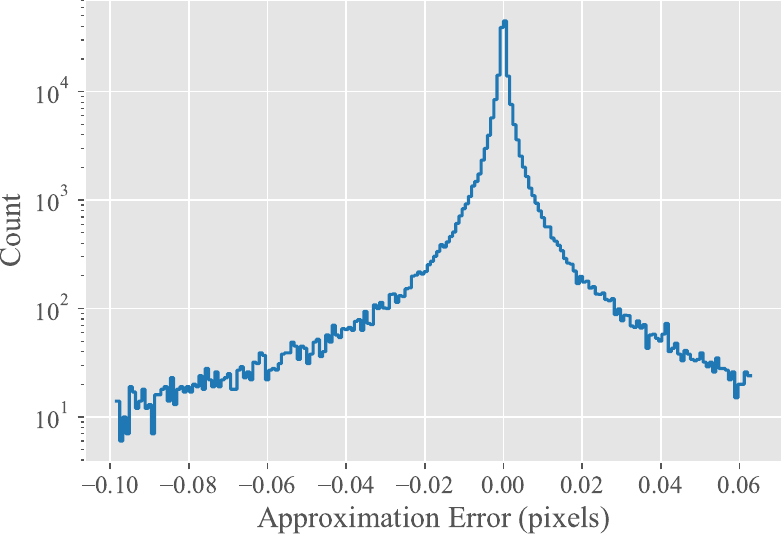}
    \caption{\textbf{Histogram over Approximation Error.} The graph shows the distribution of differences between the true and approximated residuals, \ie~$\varepsilon_S-\varepsilon_a$, scaled with focal length. Statistic taken over all clusters from all image pairs in MegaDepth-1500. }
    \label{fig:approximation_distribution}
\end{figure}

In \cref{sec:approx}, we introduced a proxy residual to approximate the sum of squared Sampson residuals for a cluster.
In this section, we experimentally validate this approximation and show that it provides a tight approximation of the true residual.
We use 128 clusters per image pair obtained via Kmeans-9D, as defined in the previous section.
For each cluster $\mathcal{C}$, we compute the per-cluster average Sampson residuals exactly as
\begin{equation}
    \varepsilon_S^2 = \frac{1}{|\mathcal{C}|} \sum_{(\xx,\xxb)\in\mathcal{C}} \mathcal{E}(E_{gt},\xx, \xxb),
\end{equation}
and the squared approximate cluster residual as
\begin{equation}
    \varepsilon_a^2 =  \frac{1}{|\mathcal{C}|} \frac{\|M\text{vec}(E_{gt})\|^2}{\|(E_{gt})_{12}\cc\|^2 + \|(E_{gt}^T)_{12}\ccb\|^2} \enspace .
\end{equation}
In Figure~\ref{fig:approximation_distribution} we show the distribution of the difference $\varepsilon_S - \varepsilon_a$ between the true residual and the approximation, computed on the 1500 image pairs from MegaDepth-1500 with \num{10000} DKM matches.
The residuals are evaluated for the ground truth essential matrix $E_{gt} = [\vec{t}_{gt}]_\times R_{gt}$.
Note that positive values indicate that the approximation gives too small residuals, while negative values correspond to the approximation giving too large residuals. We plot residuals between the 1\textsuperscript{st} and 99\textsuperscript{th} percentile. Thus we see that more than 98 \% of the residuals have an absolute approximation error of less than 0.1 pixels, with a small bias towards getting too large residuals.

In Table~\ref{tbl:approx_inlier_runtime} we compare the inlier ratios obtained through exact and approximate residuals, using the ground-truth pose. We compute average inlier ratios for each of the 1500 image pairs in MegaDepth-1500.
Note that we compute the ratio of inlier \textit{clusters} for $\varepsilon_a$, while for the exact residual $\varepsilon_S$ the inlier ratio is computed per \textit{match}.
We also include the runtime for computing the two residuals for an image pair (equivalent to scoring a model in RANSAC).
We see that the approximated residual gives slightly lower inlier ratios, which decrease slowly for coarser clustering, while the speedup in computing time is significant.

\begin{table}[ht]
\centering
\resizebox{\columnwidth}{!}{
\begin{tabular}{l c c l c c} \toprule
    & \multicolumn{2}{c}{Inlier ratio} && \multicolumn{2}{c}{Runtime ($\mu$s)} \\ \cmidrule{2-3} \cmidrule{5-6}
    Type of residual & Med. & Avg. &&  Avg. & Speedup\\ \midrule
    Exact  & 0.87 & 0.82 && 155.8  &  1.0x \\
    Approx. $K=1024$ & 0.85 & 0.80 && 31.1 &  5.0x \\
    Approx. $K=512$ & 0.84 & 0.80 &&  14.8  &  10.5x \\ 
    Approx. $K=256$ & 0.84 & 0.79 &&  7.1 & 21.9x \\ 
    Approx. $K=128$ & 0.82 & 0.78 && 3.2 &  48.7x \\ 
    Approx. $K=64$ & 0.80 & 0.76&&  1.3 &  119.8x \\ \bottomrule
    \end{tabular}
}
\caption{\textbf{Comparison of RANSAC Scoring.} We compare the inlier ratios when calculating exact and approximate residuals for different numbers of clusters $K$, using the ground-truth essential matrix on all image pairs in MegaDepth-1500. The average and median ratios over all image pairs is reported. Note that inlier ratios are calculated per match for the exact residuals, and per cluster for approximate residuals. The runtime shows the average cost of scoring a model with MSAC in RANSAC.}
\label{tbl:approx_inlier_runtime}
\end{table}

\subsection{Ablation on Integration in RANSAC}\label{sec:ablation_ransac}
In this section, we explore different ways of integrating our method in robust estimation with RANSAC. Robust estimation is typically done in two stages: the hypothesize-and-verify loop, which alternates between estimating and scoring models, and model refinement, which is done on the best model after the stop criterion is reached.

We evaluate the effect of our summarization scheme both on the model scoring and refinement stages of the estimation.
For both stages, we evaluate usage of the representative match residuals~\eqref{eq:cost_cluster}, and the approximate residuals~\eqref{eq:cost_approx}. We will refer to these methods as \textit{Center} and \textit{Approximate}, respectively.
For the refinement stage, we also evaluate usage of the original residuals~\eqref{eq:original_cost}, referred to as the \textit{Dense} method.
With two model scoring costs to evaluate (\textit{Center} and \textit{Approximate}) and three refinement costs (\textit{Center}, \textit{Approximate}, and \textit{Dense}) this results in 6 possible combinations for our ablation study, in addition to the fully dense baseline. However, we omit \textit{Approximate} scoring followed by \textit{Center} refinement, since this would use less information in the refinement than what was used for the model scoring.

For all of the above methods, we perform model estimation using the 5-point solver~\cite{nister2004efficient}, sampling from the representative matches. Note that this sampling has no effect on runtime compared to sampling from the dense matches, since convergence of RANSAC is only based on the scoring.
Since we get up to 9 constraints from each $M_k$ matrix, it is in principle also possible to estimate a model from a single summarized correspondence.
However, we found that this generally gives worse performance, see suppl.~material.

In~\cref{tab:ablation}, we show AUC@$5^\circ$ and median runtime for the different combinations of model scoring and refinement costs,  compared with a fully dense baseline. We report the average and standard deviation of the AUC for 10 runs using different seeds.
In the table, we abbreviate the methods using a three letter combination, where the first letter denotes sampling method, the second letter denotes scoring method, and the third letter denotes refinement method.
We see that scoring using \textit{Approximate} or \textit{Center} residuals are both significantly faster than \textit{Dense}. 
\textit{Center} scoring is the fastest while not significantly less accurate than \textit{Approximate}.
For refinement, however, \textit{Approximate} residuals give a small but significant improvement in score over \textit{Center}, but at an increased runtime. This gives us a possible trade-off, where we can sacrifice some runtime for better accuracy.

\begin{figure}
    \centering
    \includegraphics[width=\linewidth]{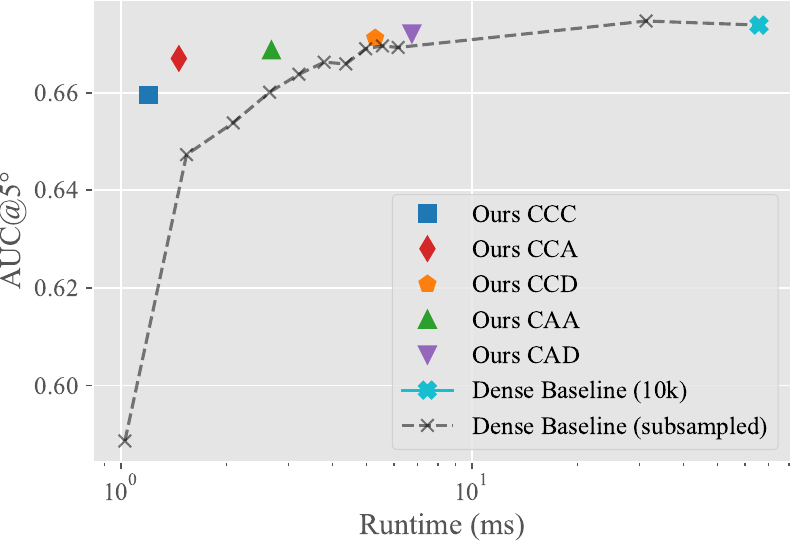}
    \caption{\textbf{AUC-Runtime Trade-off in Ablation Study.} The methods in~\cref{tab:ablation} are plotted together with dense baseline for different samplings from DKM.}
    \label{fig:ablation}
    
\end{figure}

We also compare our ablated methods with a dense baseline on subsampled DKM-matches, obtained by 
querying the balanced DKM sampler for a lower number of correspondences.
In~\cref{fig:ablation}, we plot AUC@$5^\circ$ vs. runtime for our methods and different number of DKM matches.
We see that compared to subsampling, most of our methods give shorter runtime for the same performance; in particular CCC and CCA.
This indicates that our summarization better preserves the geometric constraints, compared to reducing the number of correspondences extracted from the matcher.

\begin{table}[ht]
    \centering
    \begin{tabular}{l c cc}
    \toprule
    Method & AUC@$5^\circ$ & RT (ms) & Speedup \\
    \midrule
    
     DDD (baseline) & \tabnum{67.38}{0.10} & 66.0 & 1.0x \\

    CAD & \tabnum{67.21}{0.08} & 6.8 & 9.8x \\
    CCD & \tabnum{67.12}{0.08} & 5.3 & 12.3x \\
    CAA & \tabnum{66.87}{0.07} & 2.7 & 24.6x \\
    CCA & \tabnum{66.70}{0.06} & 1.5 & 45.2x \\
    CCC & \tabnum{65.95}{0.10} & \best{1.2} & \best{55.0x} \\

    \bottomrule
\end{tabular}

    \caption{
    \textbf{Ablation on RANSAC Integration.} We compare usage of our two summarization schemes in different stages of the estimation. 
    Method names denote the what data is used in sampling, scoring, and refinement. 
    For example, ``CAD" means sampling from representative matches (C), scoring with the summarized approximation (A), and refinement with the dense matches (D).
    }
    \label{tab:ablation}
\end{table}

\begin{table*}
    \centering
    
\scriptsize
\begin{tabular}{ll ccccc ccccc}
    \toprule
    & &\multicolumn{5}{c}{MegaDepth-1500} &\multicolumn{5}{c}{ScanNet-1500}  \\
    \cmidrule(r){3-7} \cmidrule(l){8-12}
    Matches & Estimator 
        & AUC@$5^\circ$ & AUC@$10^\circ$ & AUC@$20^\circ$ & RT (ms) & Speedup 
        & AUC@$5^\circ$ & AUC@$10^\circ$ & AUC@$20^\circ$ & RT (ms) & Speedup \\
    \midrule

    % DKM
    \multirow{3}{*}{DKM} & Dense
        % MegaDepth
        & 67.4  &  79.6  &  88.0  &  66.6 & 1.0x 
        % ScanNet
        & 31.3 & 52.6 & 69.9 & 68.8 &  1.0x \\
    & Ours CCA 
        % MegaDepth
        & 66.7 & 79.1 & 87.5 & 1.5 & 45.2x 
        % ScanNet
        & 30.9 & 52.1 & 69.6 & 1.5 & 46.7x \\
    & Ours CCC 
        % MegaDepth
        & 66.0 & 78.7 & 87.3 & 1.2 & \best{55.0x} 
        % ScanNet
        & 30.4 & 51.7 & 69.3 & 1.2 & \best{58.6x} \\
    \midrule

    % RoMA
    \multirow{3}{*}{RoMA} & Dense 
        % MegaDepth
        & 70.1 & 81.5 & 89.3 & 60.6 & 1.0x
        % ScanNet
        & 33.3 & 55.2 & 72.2 & 64.6 & 1.0x \\
    & Ours CCA 
        % MegaDepth
        & 70.1 & 81.6 & 89.4 & 1.4 & 42.2x
        % ScanNet
        & 33.0 & 54.9 & 72.0 & 1.4 & 46.3x \\
    & Ours CCC 
        % MegaDepth
        & 69.3 & 81.1 & 89.2 & 1.2 & \best{52.1x}
        % ScanNet
        & 32.8 & 54.7 & 71.8 & 1.1 & \best{58.3x} \\
    \midrule

    % ASpanFormer
    \multirow{3}{*}{\makecell{ASpan-\\Former}} & Dense 
        % MegaDepth
        & 66.3 & 78.8 & 87.5 & 14.4 & 1.0x
        % ScanNet
        & 30.3 & 50.5 & 66.5 & 15.2 & 1.0x\\
    & Ours CCA 
        % MegaDepth
        & 65.8 & 78.5 & 87.4 & 1.5 & 9.6x
        % ScanNet
        & 29.7 & 49.9 & 66.1 & 1.5 & 10.3x \\
    & Ours CCC 
        % MegaDepth
        & 64.1 & 77.4 & 86.8 & 1.2 & \best{12.1x}
        % ScanNet
        & 28.9 & 49.2 & 65.7 & 1.2 & \best{13.0x} \\
    % SP+LG
    \midrule
    SP+LG & Dense 
        % MegaDepth
        & 63.7 & 77.2 & 86.5 & 5.0 & N/A
        % ScanNet
        & 21.8 & 39.6 & 55.6 & 3.0 & N/A \\

    \bottomrule
\end{tabular}

    \caption{
    \textbf{Comparison with State-of-the-art.} We compare our method with a dense baseline for several dense matchers on both MegaDepth-1500 and ScanNet-1500. For comparison, we also include AUC and runtime for state-of-the-art sparse matches (SP+LG).
    }
    \label{tab:sota_comparison}
\end{table*}

\subsection{Comparative Evaluation in RANSAC} \label{sec:final_comparison}

We evaluate our match summarization scheme on robust estimation using different dense matchers.
We run both the CCC- and CCA-methods, since they represent a possible trade-off between accuracy and runtime.
We compare with a fully dense baseline  on the full set of matches.
To show that our results do not depend on the dense matcher, we report results using DKM~\cite{edstedt2023dkm}, RoMA~\cite{edstedt2024roma}, and ASpanFormer~\cite{chen2022aspanformer}.
From DKM and RoMA, we always sample \num{10000} matches, while ASpanFormer gives a variable amount of matches. For image pairs where ASpanFormer gives less than 128 matches, we fall back to using dense estimation.
The average results from 10 different seeds are shown in~\cref{tab:sota_comparison}.
The reported runtime is the median over all image pairs.
For the convenience of the reader, we have also included estimation from sparse SuperPoint~\cite{detone2018superpoint} + LightGlue~\cite{lindenberger2023lightglue} keypoints.
In summary, we see that our summarization scheme achieves comparable accuracy to the dense baseline at a fraction of the runtime, irrespective of the dense keypoint matcher used.

Finally, we present an experiment on the WxBS benchmark~\cite{mishkin2015wxbs}.
This benchmark contains extremely challenging wide-baseline image pairs, where GT-correspondences are provided.
As it does not provide intrinsics, we evaluate fundamental matrix estimation.
The metrics are computed by checking consistency of the estimated $F$ with the GT matches.
For comparison, we also run on dense matches extracted with MASt3R~\cite{leroy2024master}, where we disabled subsampling in the fast reciprocal matching.
The results in \cref{tab:wxbs} are qualitatively similar to the calibrated setting, showing that our method significantly speeds up the robust estimation with a marginal loss in accuracy.

\begin{table}
    \centering
    \scriptsize
    \begin{tabular}{llccc}
    \toprule
    Matches & Estimator & Recall (10px)$\uparrow$ & RT (ms)$\downarrow$ & Speedup$\uparrow$ \\
    \midrule
    \multirow{3}{*}{RoMa}
    & Dense    & 86.3 & 63.0 & 1.0x \\
    & Ours CCA & 85.1 & 1.45 & 43.3x \\
    & Ours CCC & 83.8 & \best{0.82} & \best{77.0x} \\
    \midrule
    \multirow{3}{*}{MASt3R}
    & Dense    & 63.4 & 37.1 & 1.0x \\
    & Ours CCA & 63.2 & 1.43 & 25.8x  \\
    & Ours CCC & 61.6 & \best{0.76} & \best{49.6x} \\
    \bottomrule
\end{tabular}

    \caption{\textbf{Fundamental Matrix Estimation on WxBS.}}
    \label{tab:wxbs}
\end{table}

\section{Conclusion}
\label{sec:conclusion}

In this work, we have shown that dense keypoint matches, while they improve estimation compared to sparse matches, contain significant redundancies.
This redundancy makes it possible to heavily subsample the matches while keeping most of the geometric constraints.
We demonstrated a method to significantly speed up geometric estimation with little effect on the estimation accuracy.
We additionally showed that this reduction in accuracy can be partially recovered by using our proposed summarized meta-correspondences.
With this scheme, we were able to reduce the number of residuals per image pair from the number of matches $N$, to $9K$ where $K$ is the number of clusters -- a hyperparameter we can choose such that $N \ll 9K$.
We also get a potential compression ratio in terms of storage, since the constraints from each cluster is entirely contained in a $9\times 9$-matrix, plus a single correspondence.
A limitation is that the method only makes sense if the original number of matches is very large.
Another limitation of the paper is that we have only focused on speeding up the robust estimation step, while a large part of the total runtime comes from the dense matcher itself.

{\small
\noindent\textbf{Acknowledgments.}
The project was supported by ELLIIT and
the Swedish Research Council (Grant No. 2023-05424).
}

{
    \small
    \bibliographystyle{ieeenat_fullname}
    \bibliography{main}
}

\twocolumn[{%
    \renewcommand\twocolumn[1][]{#1}%
    \maketitlesupplementary
    \vspace{2em}
    \parbox{\linewidth}{
        \normalsize
        \centering
        \begin{tabular}{lll c ccc c ccc}
    \toprule
    & & && \multicolumn{3}{c}{AUC} && \multicolumn{3}{c}{Runtime (ms)}\\
    \cmidrule{5-7} \cmidrule{9-11}
    Sampling & Scoring & Refinement && $5^\circ$ & $10^\circ$ & $20^\circ$ && Med. & Avg. & Speedup \\
    \midrule
    
     Dense
     & Dense & Dense && \tabnum{67.38}{0.10} & \tabnum{79.61}{0.08} & \tabnum{87.96}{0.06} && 66.0 & 79.0 & 1.0x \\

    \midrule
    \multirow{5}{*}{Center}
    & Approx. & Dense && \tabnum{67.21}{0.08} & \tabnum{79.65}{0.06} & \tabnum{88.12}{0.05} && 6.8 & 9.3 & 9.8x \\
    & Center & Dense && \tabnum{67.12}{0.08} & \tabnum{79.36}{0.08} & \tabnum{87.68}{0.09} && 5.3 & 6.9 & 12.3x \\
    & Approx. & Approx. && \tabnum{66.87}{0.07} & \tabnum{79.43}{0.05} & \tabnum{87.98}{0.05} && 2.7 & 4.4 & 24.6x \\
    & Center & Approx. && \tabnum{66.70}{0.06} & \tabnum{79.09}{0.08} & \tabnum{87.55}{0.08} && 1.5 & 2.2 & 45.2x \\
    & Center & Center && \tabnum{65.95}{0.10} & \tabnum{78.75}{0.07} & \tabnum{87.33}{0.09} && \best{1.2} & \best{1.9} & \best{55.0x} \\

     \midrule
     \multirow{2}{*}{Approx.\textsuperscript{*}}
      & Approx. & Dense && \tabnum{64.70}{0.00} & \tabnum{76.79}{0.00} & \tabnum{85.08}{0.00} && 8.4 & 9.7 & 7.9x \\
      & Approx. & Approx. && \tabnum{63.14}{0.00} & \tabnum{75.27}{0.00} & \tabnum{83.76}{0.00} && 4.2 & 4.3 & 15.8x \\
    \bottomrule
\end{tabular}

        \captionof{table}{%
            \textbf{Ablation on Integration in RANSAC.} We compare usage of our two summarization schemes (Center, Approx.) in different stages of RANSAC. \textsuperscript{*}For sampling and model estimation with approximated sampling, we modify RANSAC to sample exhaustively.
        \vspace{2em}}
    \label{tab:ablation_extended}
    }
}]

\section{Additional Ablation Results}
In~\cref{tab:ablation_extended}, we include some additional results on the ablation on RANSAC integration to what was presented in the main paper.
In addition to AUC@5$^\circ$, we also present AUC@10$^\circ$ and AUC@20$^\circ$.
In addition to the median estimation runtime, we also present the average runtime per image pair and seed.

We also include sampling and model estimation from the summarized correspondences, using the $M_k$ matrix.
Using SVD, we find a 4-dimensional approximate nullspace to $M_k$ and use as basis in the 5-point solver.
Since this allows us to estimate a model from a single summarized correspondence, reducing the number of unique minimal samples to $K$, we modify RANSAC to sample exhaustively instead of randomly for this method.
We refer to this method as \textit{Approximated sampling}.

\end{document}